\documentclass[letterpaper, 10 pt, conference]{ieeeconf}  

\IEEEoverridecommandlockouts                              

\usepackage{amsmath, amsfonts}
\usepackage{amssymb}
\usepackage{bm} 

\usepackage{cite}
\usepackage{hyperref}
\usepackage{graphicx}
\usepackage{siunitx}
\usepackage{xcolor}

\usepackage{placeins}
\usepackage{mathtools}

\usepackage{booktabs}

\usepackage{tabularx}
\usepackage{array}     
\newcolumntype{C}{>{\centering\arraybackslash}X}

\usepackage{booktabs}
\usepackage[table]{xcolor}
\usepackage{tikz}
\usetikzlibrary{arrows.meta, positioning}
\usepackage{printlen}

\usepackage{censor}

\newcommand{\censorNormal}[1]{%
    \begingroup
    \censor{#1}%
    \endgroup}
    
\newcommand{\censorFootnote}[1]{%
    \begingroup
    \setlength{\censorruleheight}{1.5ex}%
    \setlength{\censorruledepth}{0ex}%
    \blackout{#1}%
    \endgroup}

\StopCensoring

\newcommand{\vect}[1]{\ensuremath{\bm{#1}}}
\def \q {{\vect{q}}}
\def \F {{\vect{F}}}
\def \z {{\vect{z}}}
\def \u {{\vect{u}}}

\def \x {{\vect{x}}}

\DeclarePairedDelimiter{\abs}{\lvert}{\rvert}

\title{\LARGE \bf
Grasping by interconnection: \\
robust  closing motions from coarse object templates
}

\author{
    \censorNormal{Julien Vanderheyden}$^1$, 
    \censorNormal{Guillaume Drion}$^1$,
    \censorNormal{Fulvio Forni}$^{2}$, 
    \censorNormal{Pierre Sacré}$^{1}$ 
    \thanks{This work was supported by \censorFootnote{the Belgian Government through the Federal Public Service Policy and Support}.}
    \thanks{$^{1}$\censorFootnote{J.\ Vanderheyden}, \censorFootnote{G.\ Drion}, and \censorFootnote{P.\ Sacré} are with the \censorFootnote{Department of Electrical Engineering and Computer Science, University of Liège, Belgium (julien.vanderheyden@uliege.be; gdrion@uliege.be; p.sacre@uliege.be)}.}
    \thanks{$^{2}$\censorFootnote{F.\ Forni} is with the \censorFootnote{Department of Engineering, University of Cambridge, United Kingdom (f.forni@eng.cam.ac.uk)}.} 
}

\begin{document}

\maketitle
\thispagestyle{empty}
\pagestyle{empty}

\begin{abstract}


Dexterous robot hands must often grasp objects whose shape, size, and pose are known only approximately.
Grasp planners typically require accurate object models or correct errors with feedback, but how much inaccuracy a closing motion can tolerate on its own remains unclear.
To address this question, we designed a motion planner based on four principles: a coarse template of the object, human grasp types, an object-centric interaction, and compliant, sliding contacts instead of prescribed contact points.
This paper presents the planner, implemented through virtual model control, and its evaluation on a Shadow Dexterous Hand.
Without feedback, the planned closing motions tolerated size errors of about \qty{1}{cm} and pose errors of several centimeters and tens of degrees, a wider range than a state-of-the-art data-driven planner in 25 of 27 tested conditions.
They also grasped \qty{82.5}{\percent} of 80~everyday objects and succeeded within an autonomous pipeline.
Robustness can thus be designed into the closing motion itself, rather than left only to feedback.
This planner opens a path toward reliable manipulation in uncertain settings, which we will pursue by combining it with adaptive feedback control on the physical hand.

\end{abstract}

\section{Introduction}
\label{sec:intro}

Dexterous grasping in unstructured environments remains an open challenge in robotics~\cite{billard2019}. 
The distributed, discontinuous contacts between the hand and the object often lead to failures, which planners mitigate with  detailed information about the object alongside reliable, accurate sensing. 
In practice, however, object details are uncertain, and visual and tactile sensing are noisy. 
Reported pose estimation errors reach several centimeters in translation and several tens of degrees in orientation~\cite{nguyen2025bop}, the same magnitude as the spacing between the fingers. At this level of error, a planner cannot  determine which part of the object each finger will touch.

Current approaches to grasping typically rely on accurate object models, obtained either from a library, which limits generalization, or from sensor-based reconstruction, which introduces its own challenges. 
Visual perception, for example, is often compromised by clutter, occlusions, and changing lighting, which lead to noisy or incomplete object estimates. Other approaches reduce the need for accurate models by relying on continuous corrections based on   tactile or force feedback~\cite{dang2014stable}. However, this makes it difficult to assess whether the planned motion was effective: is successful grasping the result of good motion planning or good feedback adaptation and control? 
At the fundamental level, these approaches hide the robustness problem inherent in grasping: how much inaccuracy in the object description can a closing motion tolerate on its own, without additional sensing or feedback correction?

\begin{figure}[t]
    \centering
    \vspace*{3mm}
    \includegraphics[width= 0.85\columnwidth]{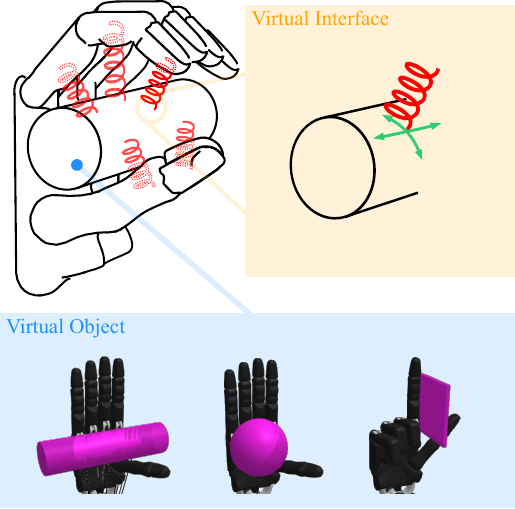}
    \caption{%
    Grasping by interconnection: the closing motion results from coupling the hand to a coarse template of the object, not from a prescribed trajectory. Each template (cylinder, sphere, or flat box) defines a \emph{virtual object} (bottom). A \emph{virtual interface} couples the virtual hand to this object through virtual springs and dampers (top); their terminals slide on the template surface (green arrows), so no contact point is prescribed. The motion is generated by virtual model control (VMC).
    } 
    \vspace{-4mm}
    \label{fig:method_overview}
\end{figure}

To answer this question, we propose a new motion planner that starts from a coarse description of the object to generate the closing motion of each finger.
This planner is based on four principles.
First, accurate object estimation is not realistic, so the planner starts from a \emph{coarse template} of the object: a cylinder, a sphere, or a flat box, with its pose and size.
Second, the motion of an anthropomorphic hand should take advantage of biological evolution, so the planner reuses \emph{human grasps}: each template is paired with a grasp type from the human grasp taxonomy~\cite{feix2015, bullock2013}, and the fingers coordinate following human motor behavior.
Third, grasping should be object-centric, so the planner is based on an \emph{object-centric interaction} between the hand and the object rather than assigning a target position to each finger.
Fourth, compliance plays a major role in managing contacts, so the planner relies on simulated \emph{compliant contacts}: contact points are not prescribed but derived from simulated physical interactions.


The motion of each finger is then planned through a \emph{virtual mechanical interface} between a virtual representation of the robot hand and the coarse templates. This approach is inspired by classical virtual model control (VMC)~\cite{prattVirtualModelControl1995,pratt2001virtual, larby2026, zhang2024}.
Coupling a hand to a template amounts to connecting mechanical elements, and passivity keeps the interaction stable without requiring exact contact models. This approach suits our principles: it is inherently object-centric and compliant, human grasp taxonomy is enforced through the mechanical structure of the virtual interface, and the contact points of the fingers are free to move on the surface of the object template, and are therefore not predefined.
As illustrated in Fig.~\ref{fig:method_overview}, the virtual interface generates virtual \emph{forces} that drive the motion of the virtual fingers. The closing motion  is thus planned by  simulation. We call this approach \emph{grasping by interconnection}. 
Once the planned motion is computed, the physical hand tracks it. No measurements of the real object are used during closure.
This paper presents the planner and the evidence for its robustness.
Section~\ref{sec:relatedwork} contextualizes the planner within the wider landscape of grasp planning.
Section~\ref{sec:method} describes how it implements the four principles.
Section~\ref{sec:experiments} reports experiments on a Shadow Dexterous Hand~\cite{kochan2005shadow}, covering errors in object size and pose, variations of the stiffness and damping parameters, everyday objects, and a comparison with a state-of-the-art data-driven planner.
The planned motions tolerated size errors of about \qty{1}{cm} and pose errors of several centimeters and tens of degrees, a wider range than $\mathcal{D(R,O)}$ Grasp~\cite{wei2024dro} in 25 of 27 tested conditions, and grasped \qty{82.5}{\percent} of 80 everyday objects.

The scope of this paper is deliberately narrow. We only address the planning of the closing motion of the hand, starting from a given pre-grasp pose. Reaching, clutter, and the selection of the object template are essential to any grasping pipeline but fall beyond this scope. 
Our planner does not estimate the description inaccuracy and does not adapt to it. It reads a template and generates one closing motion. 
We view this planner as a first step toward closed-loop grasping by interconnection. 

\section{Related work}
\label{sec:relatedwork}



\subsection{Standard grasping approaches}
\label{sec:standard}
Analytical and data-driven grasping methods achieve strong performance under controlled conditions, but depend on accurate object models, which limits their applicability in uncertain environments.  

Analytical methods compute force-closure grasps by optimizing over contact geometry~\cite{ferrari1992planning,bicchi2000robotic}. These techniques inherently require detailed object models, either known \textit{a priori}, which often limits their use to a predefined set of objects~\cite{bohg2013}, or reconstructed online from partial observations, which introduces errors due to the noise of 3D sensing~\cite{lum2024get}. 
In both cases, the reliability of the planned grasp is limited by the accuracy of the object representation or reconstruction. 

Data-driven approaches learn grasp strategies from large datasets, implicitly constructing internal object representations. 
They generalize to unseen objects either by reconstructing the object shape~\cite{varley2017shape} or by inferring grasps directly from point clouds~\cite{wei2024dro}. Both strategies assume that the input data at test time matches the training distribution, which makes them sensitive to sensor noise and domain shift. 

Beyond their reliance on accurate object models, both approaches typically output a final grasp, defined by contact points or a grasp pose, and leave the motion toward it to a generic planner or controller.
When the size or pose of the object is wrong, contacts occur before or away from the predicted points, and the grasp outcome depends on a motion that was not designed to handle these errors.
Our planner mitigates these issues  through an object-centric closing motion relative to a coarse template. In contrast to predefined contact points, fingers track a coordinated closing motion adapted to the broad shape of the object.

\subsection{Object shape approximation}

An alternative to precise object reconstruction is to build approximate models and trade geometric detail for robustness~\cite{ekvall2007}. Cognitive science supports this choice: human object recognition and manipulation favor simplified geometric approximations~\cite{hoffman1984, biederman1987}.


Among the simplest strategies, shape primitives, such as cylinders, spheres, and boxes~\cite{miller2003, ekvall2007}, and box-based representations~\cite{huebner2008, huebner2008selection, palleschi2023} have been successful in finding grasp regions. 
Richer representations, such as superquadrics~\cite{goldfeder2007, vezzani2017}, medial axes~\cite{przybylski2010, przybylski2011}, and skeletal models~\cite{vahrenkamp2018}, capture more geometric detail.
However, their dependence on accurate shape estimation makes them more vulnerable to sensor noise.
This trade-off between geometric expressiveness and robustness motivates the use of \emph{coarse templates}.

Our planner differs from the methods above in how the template is used.
These methods derive a target hand posture, leaving the associated motion planning to external algorithms, such as the robot hand controller.
In contrast, we use templates to determine the entire closing motion. 
As shown in Figs.~\ref{fig:method_overview} and \ref{fig:grasp_implementations}, each coarse template is paired to a specific virtual mechanism whose terminals slide on the template surface.
The planned motion of each finger is then derived by simulation. 
This \emph{object-centric interaction} connects to the evidence from  cognitive science, adopting a coarse object geometry to shape the motion.

\subsection{Human-inspired grasp synthesis}

Human grasping has evolved to remain robust to uncertainty in perception, object geometry, and physical interaction, making it a natural source of inspiration for robust manipulation. It also provides a principled way to manage the high dimensionality of the grasping problem.

On the one hand, \emph{grasp taxonomies} organize grasps into canonical types, providing lightweight priors that guide hand configuration without a full high-dimensional optimization.  Explicit models of grasp type improve task success, even in data-driven settings~\cite{lu2019}. However, early approaches paired each grasp type with a fixed closing motion~\cite{lyons1985, stansfield1991, miller2003, romero2009}, which cannot adapt to geometric variations between objects. Methods that combine grasp types with analytical optimization~\cite{kang1997, morales2006, harada2008, vahrenkamp2018, deng2019, deng2021} still require a detailed object model to evaluate grasp quality, and methods that use grasp types in learned policies~\cite{amor2012, lundell2021, dimou2022, zhang2023} inherit the sensitivity to sensor noise discussed in Section~\ref{sec:standard}. 
The benefits of grasp taxonomies are thus often hindered by rigid or model-dependent executions.

On the other hand, \emph{hand synergies} reduce the complexity of hand control by exploiting the low-dimensional structure of natural hand motion. 
Postural synergies~\cite{santello1998, santello2016} capture the dominant modes of joint covariation in human grasping and provide a compact basis of hand postures that makes optimization and learned control tractable in real time~\cite{ciocarlie2007eigengrasps, amor2012}. However, the synergy basis is extracted from hand postures rather than from the objects being grasped: the same basis applies regardless of object shape, which leaves open how synergies cope with errors in the object description.


In our approach, we bridge these two paradigms.
We take advantage of grasp types to identify high-quality grasp configurations, which are later adapted to the variations in object size and geometry.
Rather than constraining the hand to a predefined synergy subspace, we use VMC to coordinate the fingers through an \emph{object-centric interaction}. A structured coupling  of virtual springs and dampers between the hand and the template compliantly constrains the motion of the hand, without limiting its degrees of freedom.
The taxonomy determines which coupling is appropriate for a given template, and VMC handles finger coordination.
This design aims to retain the robustness of human grasping, as verified in Section~\ref{sec:experiments}.

\section{Grasping by interconnection}
\label{sec:method}

\begin{figure}[t]
    \centering
    \vspace*{3mm}
    \includegraphics[width = 1.0\linewidth]{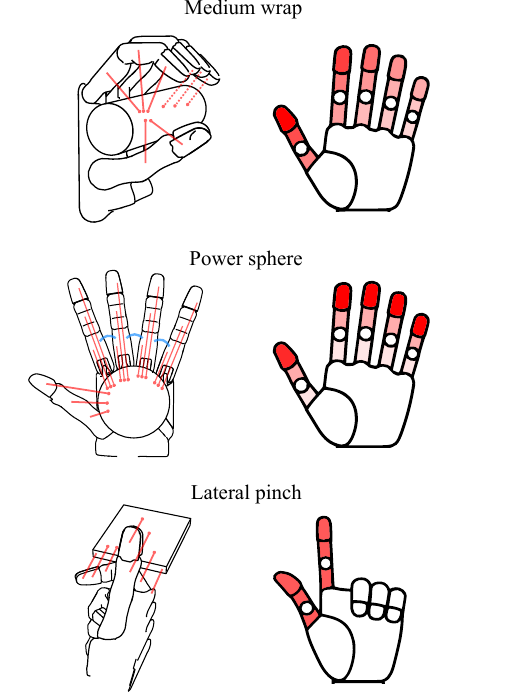}
    \caption{%
    The virtual interface encodes each grasp type through its architecture and stiffness distribution, without prescribing contact points. 
    \textit{Left.}~Contractive (red) and repulsive (blue) springs drive the fingers toward stable contact; sliding terminals (circles) move on the template surface, so contact locations result from the motion.  
    \textit{Right.}~Stiffness distribution set by $\alpha$ and $\beta$; higher color intensity indicates higher stiffness.
    }
    \label{fig:grasp_implementations}
\end{figure}

This section describes what our motion planner reads, what it computes, and how it implements the four principles. 
Section~\ref{sec:input} specifies the coarse template and its grasp type. 
Section~\ref{sec:design} explains why virtual model control implements an object-centric interaction and how the virtual hand is simulated. 
Sections~\ref{sec:virtual_object} and~\ref{sec:hand_motion} detail the two components of the planner: a virtual object, which models the template with compliant elements, and a virtual interface, which couples the virtual hand to it through sliding terminals and couplings derived from human grasps (Fig.~\ref{fig:grasp_implementations}).

\subsection{Object description}
\label{sec:input}

Objects are reduced to \emph{coarse templates}.
The planner relies on (i)~the template---a cylinder, a sphere, or a flat box---and (ii)~a small set of parameters related to the pose and the size of the template. 
The position and, for cylinders and flat boxes, the orientation of the template set the pre-grasp pose of the hand.
The closing motion itself depends only on the template size: the diameter of a cylinder or a sphere, or the thickness and width of a flat box.
A template therefore consists of a discrete class with at most eight parameters, of which at most two shape the closing motion.

The mechanical interconnection between the template and the virtual hand encodes the \emph{human grasp taxonomy}:
it pairs 
the cylinder, the sphere, and the flat box with 
the medium wrap, the power sphere, and the lateral pinch, respectively~\cite{feix2015}.
Bullock et al.~\cite{bullock2013} identified these three grasp types as the most versatile ones in household settings. Within each grasp type, the fingers' motion is coordinated following human motor behavior, as detailed in Section~\ref{sec:hand_motion}. 

No further visual, tactile, or force measurements are used. 
Following the \emph{compliant contacts} principle, the planner prescribes no contact point: contact locations are an outcome of the simulation of the virtual hand coupled to a template-based virtual mechanism, as described below.  The tolerance reported in Section~\ref{sec:experiments} is therefore a property of the planned motion, executed without feedback from the object.


\subsection{Virtual model control}
\label{sec:design}

Inspired by VMC \cite{pratt2001virtual, larby2026},
our planner frames grasping as an \emph{object-centric compliant interaction} realized by a mechanical interconnection between the virtual hand and the template object, 
rather than as a set of finger contact points. 
The associated (simulated) forces determine the planned coordinated motion of the fingers.
In the classical case of robot manipulators with revolute joints, these virtual forces $\F_i$ act along coordinates $\z_i = h_{\z_i}(\q)$ of the virtual elements, such as the extension of a spring, expressed as functions of the joint coordinates $\q$.
These forces are translated into joint torques by
\begin{equation}
    \u = \sum_{i=1}^n J_{\z_i}^T\left(\q\right) \F_i,
\end{equation}
where $J_{\z_i} = \frac{\partial h_{\z_i}(\q)}{\partial \q}$ is the corresponding Jacobian matrix. 

VMC 
guarantees that the controlled robot remains passive~\cite{Ortega2001, Spong2022, Chopra2022, Secchi2007}, within the limits of faithful approximation of the virtual mechanical elements. 
Passivity ensures stable interaction with passive objects without precise contact models, even at stiff contact transitions.  
Moreover, the use of virtual mechanism is inherently modular: grasp-specific elements can be added or reconfigured independently, which translates into highly configurable virtual interfaces for each template and grasp type.
Finally, deriving finger motion via VMC and simulation does not require inverse kinematics computation, reducing the computational burden that typically arises from the high dimensionality of anthropomorphic hands.



We emphasize that, in this paper, we use VMC as a motion planner rather than as a feedback controller, so that the closing motion is fixed before execution and its robustness can be assessed on its own.
The virtual hand is a copy of the robot hand built from virtual links, joints, masses, and inertias.
The virtual object and the virtual interface described below, which together form the virtual mechanism, are attached to it, and the resulting mechanical system is simulated.
The simulation integrates the dynamics of this virtual system,
\begin{equation}
    M(\x)\ddot{\x} + C(\x,\dot{\x})\dot{\x} + g(\x)
    = \sum_{i=1}^n J_{\z_i}^T\left(\x\right) \F_i,
\end{equation}
where $\x = (\q, \q_\mathrm{v})$ collects the joint coordinates $\q$ of the virtual hand and the coordinates $\q_\mathrm{v}$ of the virtual mechanism, namely the revolute and prismatic joints of the sliding terminals, which have low-inertia endpoints.
$M$, $C$, and $g$ are the inertia matrix, the Coriolis terms, and the gravity terms of this coupled system, and the Jacobians are taken with respect to $\x$.  
The output of the planner is the joint trajectory $\q(t)$ produced by this simulation, which the physical hand then tracks as a position reference. The simulation runs ahead of execution. 


\subsection{Virtual object}
\label{sec:virtual_object}

Each coarse template is a virtual object with a \emph{compliant surface}, whose mechanics is regulated by 
simple springs and dampers. 
The surface springs generate repulsive forces that keep the virtual fingers from penetrating the virtual object.
The force of each virtual spring is ReLU shaped
\begin{equation}
\F_\mathrm{s}
=
k_\mathrm{obj} \min\left(\abs{ \z } - z_\mathrm{rest}, 0\right)
\frac{\z}{\abs{ \z }},
\end{equation}
where $\z$, $z_\mathrm{rest}$, and $k_\mathrm{obj}$ are the extension, rest length, and stiffness of the spring, respectively. 
Each spring pushes back only when compressed below its rest length, which mimics the onset of contact.


The template size sets the geometry of the virtual object and the mechanical parameters of the virtual interface: the rest lengths $z_\mathrm{rest}$ of the repulsive springs and the placement of the sliding terminals, so that the attractive and repulsive forces drive the fingers to the surface. The stiffness and damping parameters do not depend on the size.

The dampers suppress oscillations at contact. The associated forces satisfy $ \F_\mathrm{d} = c_\mathrm{obj}(\abs{\z} )\,\dot{\z}$ with ReLU-shaped damping 
\begin{equation}
    c_\mathrm{obj}\left(\abs{\z}\right) =
    \frac{c_\mathrm{max}}{\gamma z_\mathrm{rest}} \, \max \left(\gamma z_\mathrm{rest} - \abs{\z}, 0 \right).
\end{equation}
The coefficient $\gamma=1.05$  inflates the virtual object by \qty{5}{\percent}, so that damping activates before the spring force.
All force profiles are continuous, which avoids  abrupt variations in the planned motion.  
Other passive contact models could replace these springs and dampers without changing the architecture.

\subsection{Virtual interface}
\label{sec:hand_motion}


The virtual interface couples the virtual hand to the virtual object.
Grasp synthesis combines two objectives: the identification of mechanically stable final contact configurations and the derivation of the motion trajectories that bring the hand to that configuration. In addition, different grasp types impose specific constraints that must be explicitly encoded. The three virtual interfaces of Fig.~\ref{fig:grasp_implementations} address all these aspects.

Stable contact configurations follow from compliant interaction: the terminals of the contractive springs within the virtual interface are free to move on the virtual object surface. Therefore, they attract fingers and palm toward the template without forcing predefined contact points.
These terminals slide along the template surface through prismatic and revolute joints, allowing for an entire set of final configurations, rather than a single one. 

The transient motion of the fingers toward the final, stable configuration 
is shaped by the relative stiffness and damping of the virtual elements across the whole hand.
Two complementary mechanisms, both inspired by human motor behavior, coordinate the fingers.

\begin{figure}[tbp]
    \vspace{3mm}
    \centering
    \includegraphics[width=\linewidth]{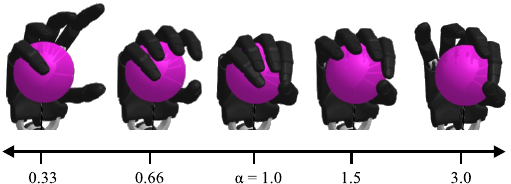}
    \caption{%
    A single parameter governs the coordination between fingers during closure. At $\alpha=1$, all fingers close symmetrically. Increasing $\alpha$ advances the thumb and index first, creating a stabilizing pinch before full-hand enclosure; decreasing $\alpha$  advances the outer fingers first.
    }
    \label{fig:sensitivity_spheres}
\end{figure}

The first mechanism  coordinates the closure of the fingers by distributing  the spring stiffnesses  according to a geometric scaling law:
\begin{equation}
    k_{ij} = k_0 \, \alpha^{\frac{i-1}{N_\mathrm{f}-1}} \, \beta^{\frac{j-1}{N_\mathrm{p}-1}},
\end{equation}
where $N_\mathrm{f}$ and $N_\mathrm{p}$ are the numbers of fingers and phalanges, $i$ is the finger index, $j$ is the phalanx index, and $k_0$ is a base stiffness chosen experimentally. 
This law reduces the stiffness design space to two parameters, $\alpha$ and $\beta$, which can be explored systematically. 
The parameter $\alpha$ sets which fingers close first, as illustrated in Fig.~\ref{fig:sensitivity_spheres}. 
The parameter $\beta$ distributes the attraction among the phalanges of each finger, which sets the closure pattern within a finger.

The second mechanism makes the fingers reach the template surface nearly simultaneously, which prevents ejection at first contact. Distance-dependent damping slows  each finger down as it approaches the surface.
Each spring is paired with a damper whose coefficient follows
\begin{equation}
    c\left(\abs{ \z } \right)
    = c_0 + c_1\, \exp\left( -\frac{\abs{ \z }}{\tau} \right).
\end{equation}
A uniform term $c_0$ suppresses oscillations, while an exponential term $c_1$ decelerates the finger near the template surface. 
The decay constant is $\tau=-d / \ln(0.2)$, where $d$ is the distance at which the exponential damping reaches \qty{20}{\percent} of its maximum value. 

The parameters of each spring and damper are adapted to the specific template and align with the human grasp taxonomy.
That is, in addition to the sliding-contact architecture, they encode the features of each grasp type (Fig.~\ref{fig:grasp_implementations}).
For the medium wrap, $\alpha > 1$ creates a stabilizing thumb--index pinch before the whole hand closes.
For the power sphere, angular springs maintain the spacing between the fingers and promote enclosure, and $\alpha = 1$ makes the fingers close symmetrically, which prevents lateral ejection (Fig.~\ref{fig:sensitivity_spheres}).
Finally, for the lateral pinch, both extremities of the index finger are driven toward the box corners, maximizing span.
Table~\ref{tab:parameters_values} lists the parameter values for each grasp type.

\begin{table}[tbp]
    \centering
    \caption{Parameter values for each grasp type.}
    \label{tab:parameters_values}
    \footnotesize 
    \setlength{\tabcolsep}{6pt} 
    \renewcommand{\arraystretch}{1.3} 
    \begin{tabularx}{\columnwidth}{@{} l *{8}{C} }
        \toprule
        & $k_\mathrm{obj}$ & $c_\mathrm{max}$ & $k_0$ & $\alpha$ & $\beta$ & $c_0$ & $c_1$ & $d$ \\
        \midrule
        Medium wrap   & 5.0 & 5.0 & 0.05 & 1.5 & 0.5 & 0.05 & 0.1 & 0.01 \\
        Power sphere & 5.0 & 5.0 & 0.05 & 1.0 & 0.1 & 0.15 & 0.2 & 0.01 \\
        Lateral pinch  & 5.0 & 5.0 & 0.1  & 1.0 & 1.0 & 0.05 & 0.1 & 0.005 \\
        \bottomrule
    \end{tabularx}
\end{table}


\section{Experiments}
\label{sec:experiments}

\subsection{Experimental setup}
\label{sec:implementation}

\begin{figure}[t]
    \vspace{4mm}
    \centering
    \includegraphics[width = 0.95\linewidth]{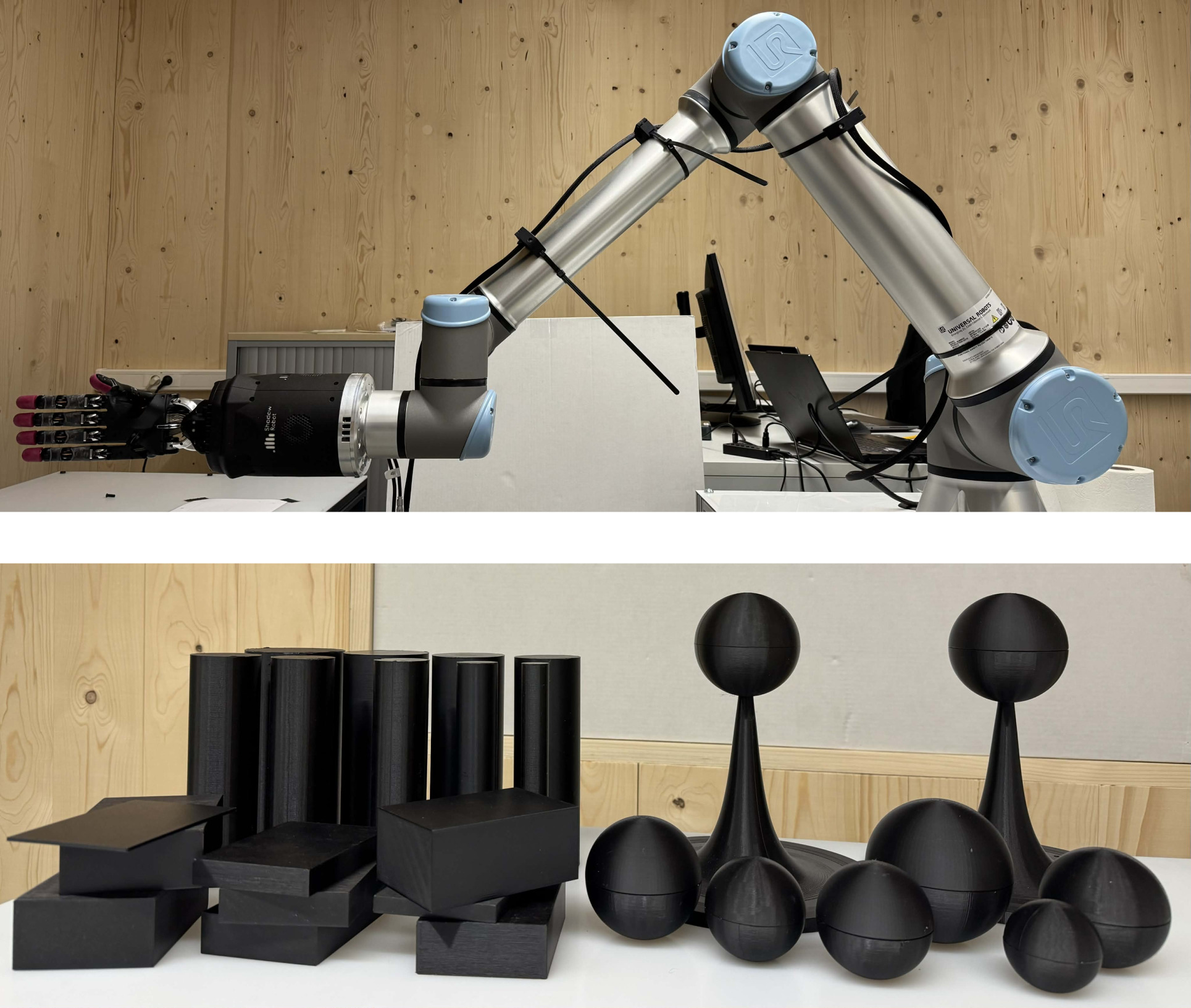}
    \caption{%
    Robustness tests use 27 3D-printed cylinders, spheres, and boxes (bottom), grasped by a Shadow Dexterous Hand on a UR10e arm (top).
    }
    \label{fig:exp_placeholder}
\end{figure}

Experiments were conducted with a Shadow Dexterous Hand~\cite{kochan2005shadow} mounted on a UR10e arm (Fig.~\ref{fig:exp_placeholder}). 
The hand is position controlled:
its joint controllers track the
closing motion computed by the planner.
All computations\footnote{The code, including the grasping pipeline of Section~\ref{sec:experiments}, will be released on a GitHub page after the double-blind review process.\label{fn:code}} use \texttt{VMRobotControl.jl}~\cite{VMRobotControl}, a Julia library for the design, simulation, control, and optimization of VMC architectures.
The test set consists of 27 3D-printed cylinders, spheres, and flat boxes covering the typical sizes of everyday objects. Cylinders were placed upright, spheres on dedicated stands, and flat boxes on elevated surfaces. 
For each grasp type, the hand started from a fixed pre-grasp pose relative to the template.
A grasp is considered successful if the object is lifted for five seconds without slipping. 

\subsection{Robustness to size and pose errors, and comparison}

\begin{figure}[tbp]
    \centering
    \vspace*{3mm}
    \includegraphics[width=1.0\columnwidth]{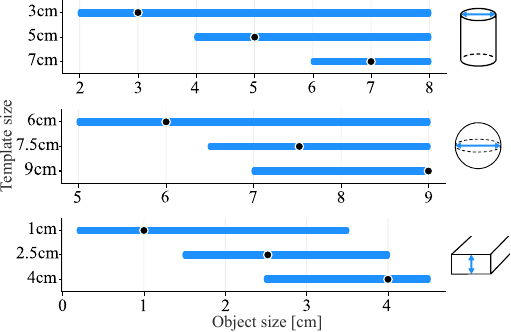}
    \caption{%
    Size errors of about \qty{1}{\centi\meter} do not compromise grasp success. Blue intervals show the range of successful grasps for each template size; sizes are diameters for cylinders and spheres and thicknesses for flat boxes; black points mark exact matches between template and object.
    }
    \label{fig:dimension_error}
\end{figure}

\begin{figure*}[t]
    \centering
    \vspace*{3mm}
    \includegraphics[width = 1.0 \linewidth]{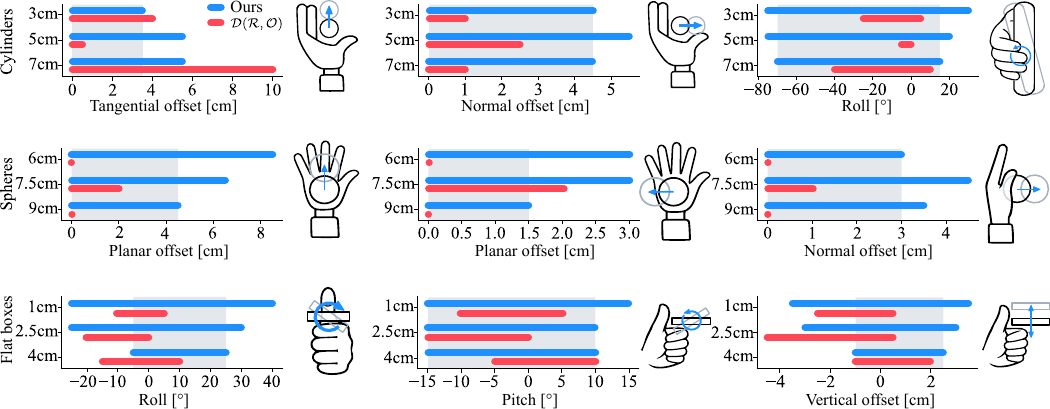} 
    \caption{%
    Our planner (blue) tolerates wider pose errors than $\mathcal{D(R,O)}$ Grasp (red) in 25 of the 27 axis--object combinations. Bars show the perturbations with successful grasps along three axes per grasp type (icons), for three object sizes; for $\mathcal{D(R,O)}$, a perturbation counts as successful if at least two of three generated grasps succeed. Grey shaded area marks the range shared by the three objects for our planner.
    }
    \label{fig:position_error}
\end{figure*}

Robustness to size errors was evaluated by choosing three template sizes for each grasp type.
The associated grasp type was applied across a range of object sizes typical of human grasping; each grasp was repeated five times, with identical outcomes.
Fig.~\ref{fig:dimension_error} shows the range of object sizes grasped with each template: errors of about \qty{1}{cm} do not induce any loss of performance.
Underestimating the size is tolerated best: the \qty{3}{cm} cylinder template grasps cylinders up to \qty{8}{cm}, the largest size tested. Although successful, these mismatched grasps are less stable and less geometry-adapted than those obtained using a matching template.

Robustness to pose errors was evaluated for each grasp type along three perturbation axes, shown by the hand
icons in Fig.~\ref{fig:position_error}. 
Increasing perturbations are applied to the three objects that match the template sizes, with results reported as blue lines. The ranges shared by the three objects (grey shaded ranges) span 1.5 to \qty{4.5}{cm} in translation and 25 to \qty{85}{\degree} in orientation.

The red lines in Fig.~\ref{fig:position_error} compare our planner with $\mathcal{D(R,O)}$ Grasp~\cite{wei2024dro}, a state-of-the-art generative grasp synthesis method that infers grasp poses from arbitrary point clouds. 
$\mathcal{D(R,O)}$ received ideal point clouds sampled from the object meshes, so both methods start from exact object information before perturbation.
Using the same perturbation protocol, three grasp poses were generated per object, and a perturbation is considered successful if at least two grasps succeed.
The hand was first moved to the nominal grasp pose and then displaced to the perturbed pose. In the case of $\mathcal{D(R,O)}$, contact during this motion may drag the object along, which reduces the actual perturbation and can lead to overestimates in the robustness of $\mathcal{D(R,O)}$. In contrast, our approach starts from an open palm, so no contact occurs. Despite this bias, 
$\mathcal{D(R,O)}$ tolerates a narrower range than our planner in 25 of the 27 axis--object combinations; in the two exceptions, tangential offsets of the 3 and \qty{7}{cm} cylinders, the drag of the object affects the estimate. We also tried to extend the comparison to DexGrasp Anything~\cite{zhong2025dexgraspanything}. However, out of the box, the produced grasps were poor, which prevented a proper comparison.


%
The gap in performance follows from the design of the two planners: 
$\mathcal{D(R,O)}$ generates static grasp poses anchored to specific contact points, which enables complex, object-specific grasps but reduces tolerance to pose errors. 
In contrast, our planner prescribes no contact points, generates the whole closing motion, and takes advantage of well-established human grasps. 
The two approaches are complementary: $\mathcal{D(R,O)}$ adapts grasps to complex shapes, whereas our planner favors robustness through its reliance on generic templates.

\subsection{Sensitivity to stiffness and damping parameters}

We evaluated the sensitivity of the planner to the parameters of the virtual interface on the same nine objects, each matching its template in size and pose. Stiffness and damping were independently varied by up to \qty{\pm 25}{\percent} from their nominal values (Table~\ref{tab:parameters_values}). 
No grasp failures occurred under damping variations, while stiffness variations caused a single failure at the boundary of the tested range, for the largest sphere. 
This failure resulted from changing $\alpha$. For the power sphere, the nominal value of $1.0$ makes the fingers close symmetrically while a deviation  disrupts the enclosure and ejects the object early, as shown in Fig.~\ref{fig:sensitivity_spheres}.
Our planner therefore does not require precise tuning within \qty{\pm 25}{\percent}: the only failure came from changing the relative stiffness across fingers, not from the absolute parameter values. 
\subsection{Everyday objects}

\begin{figure*}
    \centering
    \vspace*{3mm}
    \includegraphics[width=\linewidth]{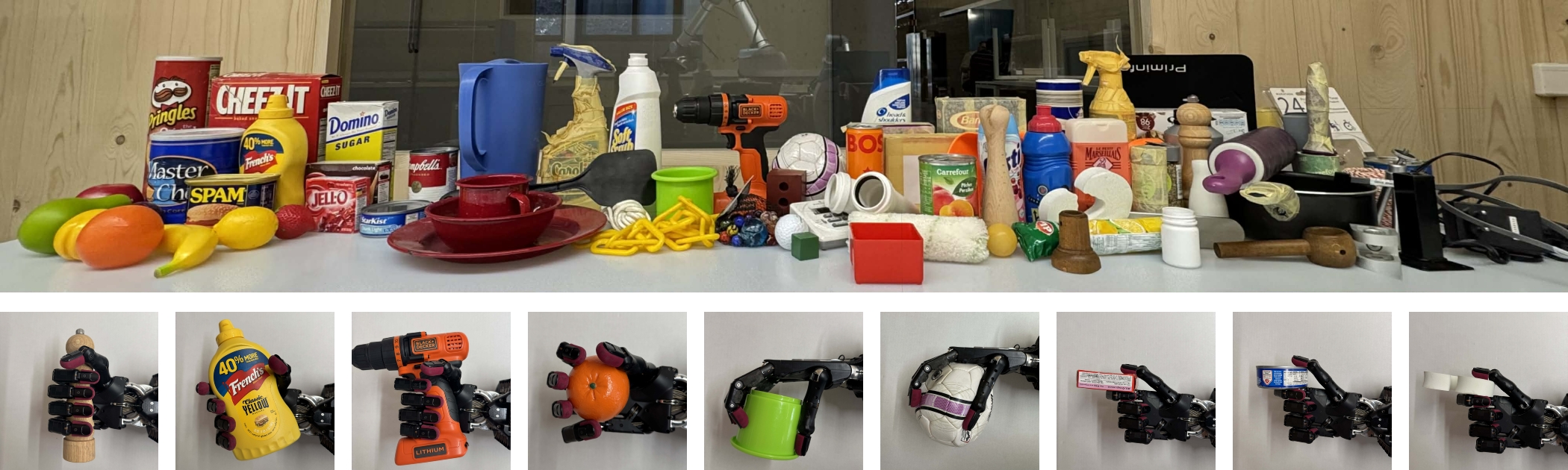}
    \caption{
        Three templates suffice to grasp \qty{82.5}{\percent} of 80 everyday objects (\qty{77}{\percent} of the YCB subset). YCB objects appear on the left, additional objects on the right; the bottom row shows example grasps.
    }
    \label{fig:ycb_objects}
\end{figure*}

We assessed the ability of our planner to grasp everyday objects. 
Each object was assigned to a coarse template that matches its dominant dimensions: a sphere for approximately isotropic objects, a cylinder for one dominant dimension, and a flat box for two. 

We evaluated this generalization on 35 objects from the YCB dataset~\cite{calli2015ycb}, primarily from the \textit{Food} and \textit{Kitchen} categories, complemented by 45 additional everyday objects (Fig.~\ref{fig:ycb_objects}). 
Our planner grasped \qty{77}{\percent} of the YCB objects and \qty{82.5}{\percent} of all 80 objects, which shows that three coarse templates approximate many everyday objects.

\begin{figure}[htbp]
\centering
\vspace*{3mm}
\includegraphics[width=0.99\columnwidth]{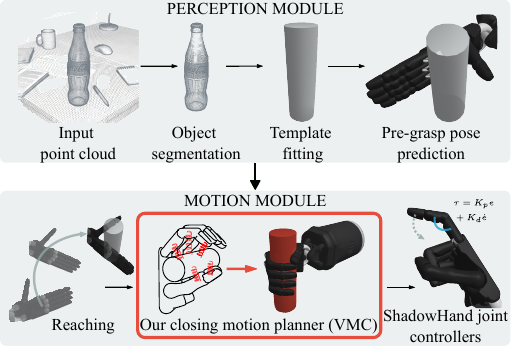}

\vspace*{3mm}

\setlength{\tabcolsep}{14pt}
\renewcommand{\arraystretch}{1.0}
\newcolumntype{V}{!{\color{black!25}\vrule width 0.4pt}}
\begin{tabular}{@{}ccc@{}}
\toprule
\footnotesize\textbf{Medium wrap} & \footnotesize\textbf{Power sphere} & \footnotesize\textbf{Lateral pinch} \\
\midrule
\includegraphics[width=0.24\columnwidth]{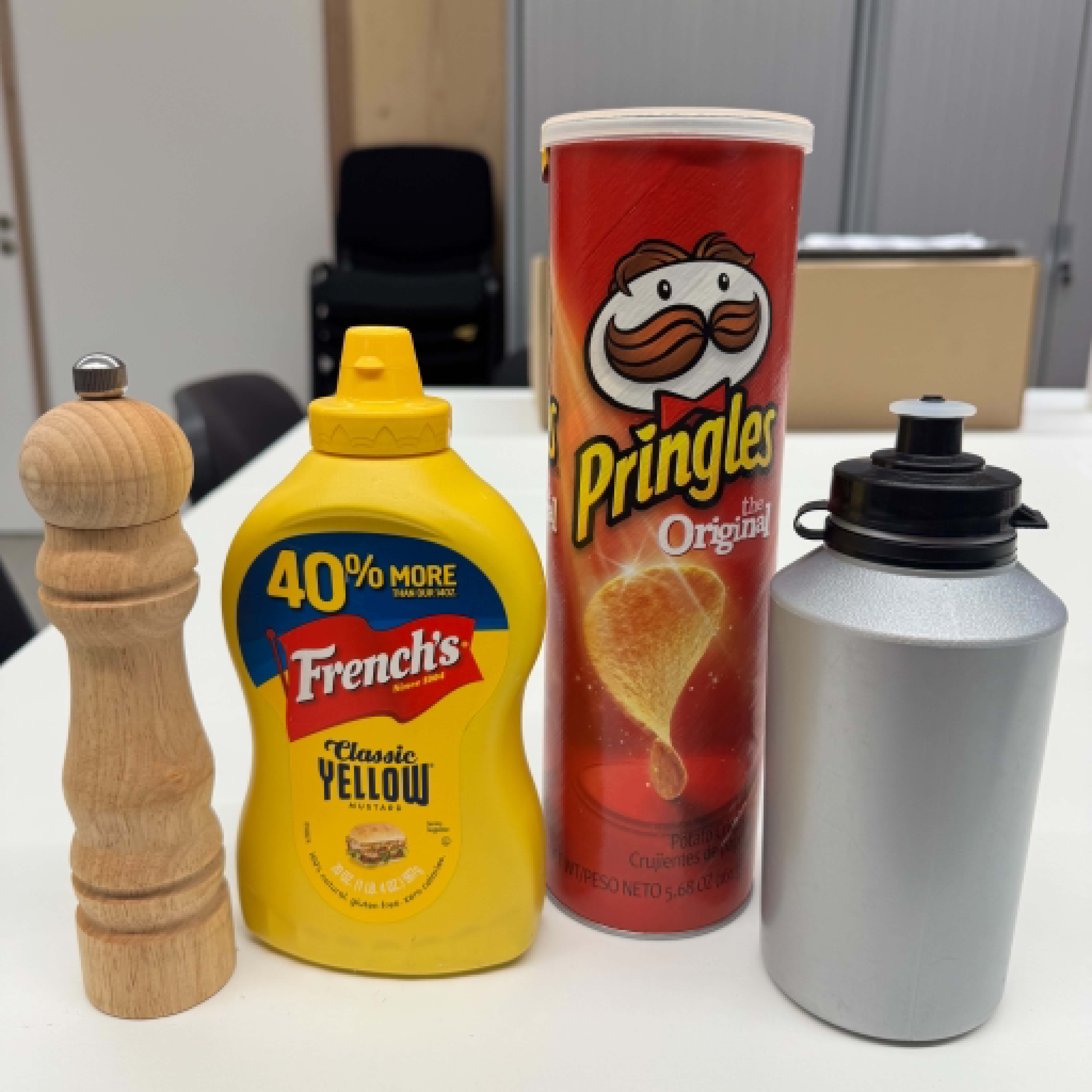} &
\includegraphics[width=0.24\columnwidth]{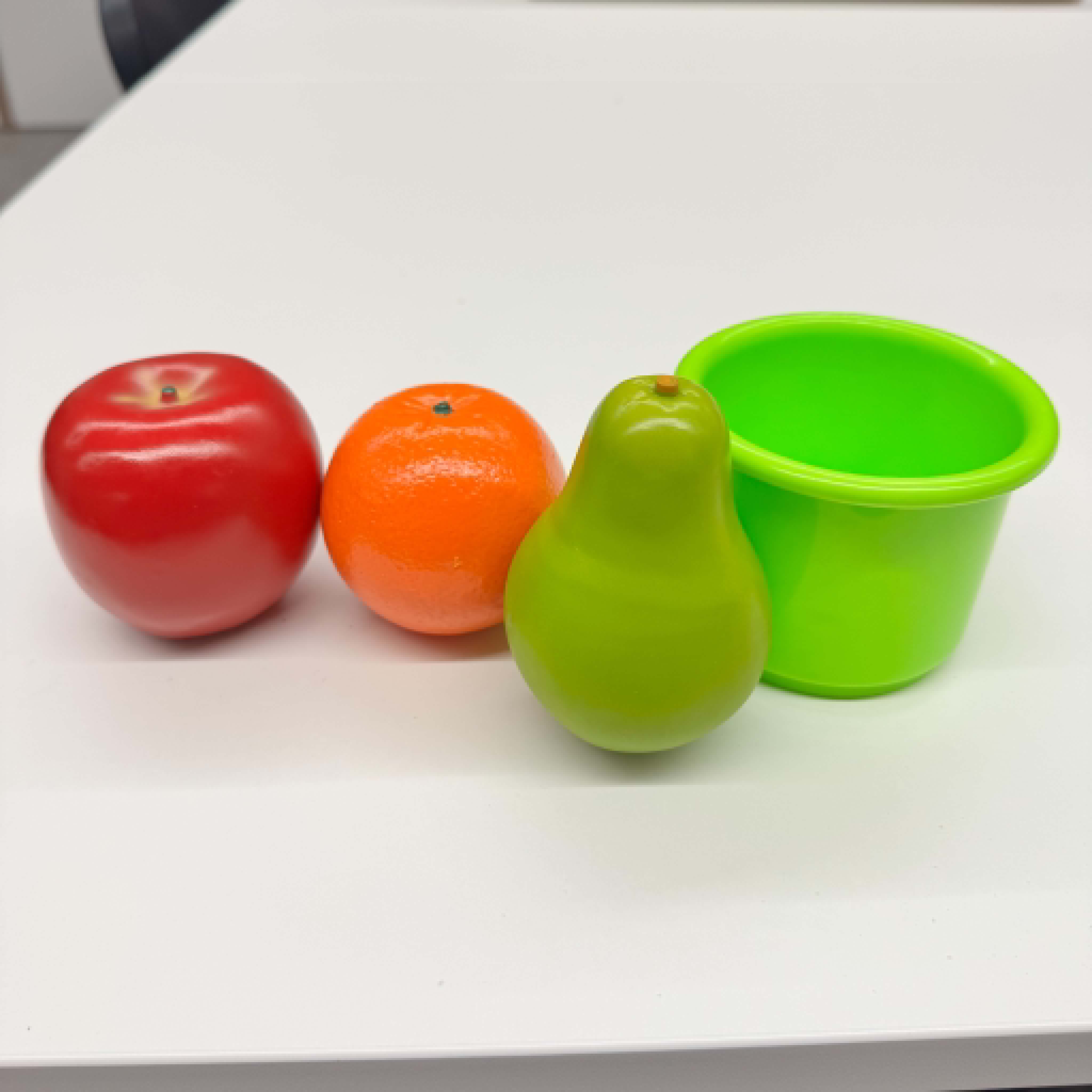} &
\includegraphics[width=0.24\columnwidth]{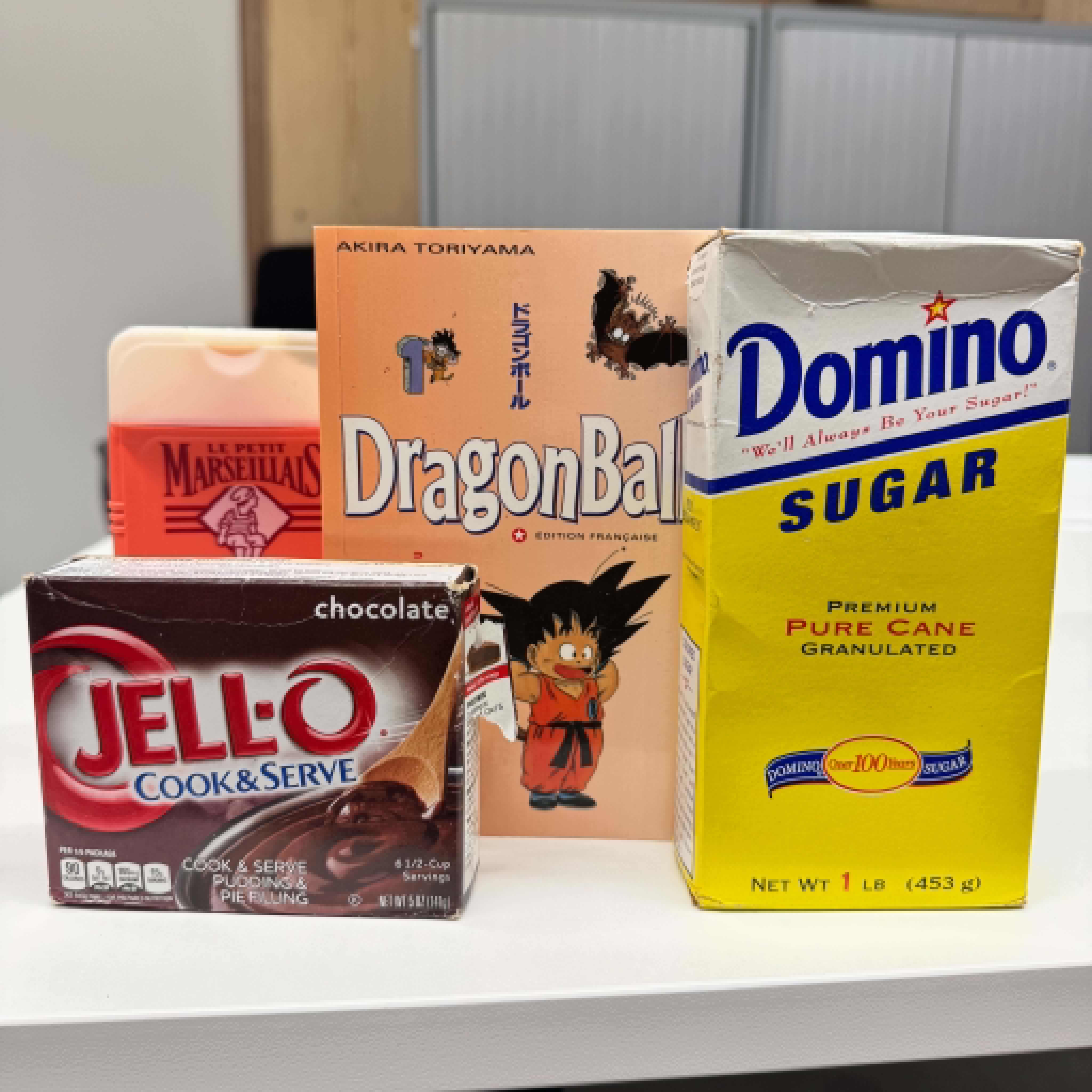} \\
\midrule
\begin{tabular}[t]{@{}l@{\hspace{6pt}}r@{}}
\scriptsize Pepper mill & \scriptsize 5/5 \\[-1.3pt]
\scriptsize Mustard & \scriptsize 5/5 \\[-1.3pt]
\scriptsize Pringles & \scriptsize 5/5 \\[-1.3pt]
\scriptsize Bottle & \scriptsize 5/5 \\
\end{tabular}
&
\begin{tabular}[t]{@{}l@{\hspace{6pt}}r@{}}
\scriptsize Apple & \scriptsize 5/5 \\[-1.3pt]
\scriptsize Orange & \scriptsize 4/5 \\[-1.3pt]
\scriptsize Pear & \scriptsize 3/5 \\[-1.3pt]
\scriptsize Green cup & \scriptsize 4/5 \\
\end{tabular}
&
\begin{tabular}[t]{@{}l@{\hspace{6pt}}r@{}}
\scriptsize Soap & \scriptsize 5/5 \\[-1.3pt]
\scriptsize Jell-O & \scriptsize 5/5 \\[-1.3pt]
\scriptsize Dragon Ball & \scriptsize 5/5 \\[-1.3pt]
\scriptsize Domino sugar & \scriptsize 3/5 \\
\end{tabular}
\\
\bottomrule
\end{tabular}
\caption{%
Within an autonomous pipeline, the planner grasps 12 everyday objects in 54 of 60 trials (\qty{90}{\percent}).
Top: pipeline modules, with our planner outlined in red.
Bottom: successes over five trials per object, grouped by grasp type.
}
\label{fig:pipeline_results}
\end{figure}

As a final validation, we integrated our planner into an autonomous grasping pipeline (Fig.~\ref{fig:pipeline_results}, top). 
An Intel RealSense D435i 3D camera provides a point cloud, from which standard lightweight methods segment the object, fit the template, predict the pre-grasp pose, and reach it; their implementation is included in the released code.
Twelve of the 80 objects were tested in five random orientations each, for a total of 60 trials.
The planner succeeded in 54 of 60 trials (\qty{90}{\percent}), which shows that it can be integrated into a conventional grasping pipeline with camera-based sensing.

\section{Conclusion}

This paper shows that the robustness of robot grasping  can be improved through proper design of the grasping motion. 
We have introduced a new motion planner rooted in four key principles and realized via VMC. 
Without relying on any real-time feedback or adaptation, the planned motions tolerated object size errors of about \qty{1}{cm} and pose errors of several centimeters and tens of degrees, wider than a state-of-the-art data-driven planner in most cases. Three coarse templates were also enough to grasp \qty{82.5}{\percent} of 80 everyday objects, so this robustness does not come at the cost of generalization.

This planner is a first step.
It provides desired trajectories which are tracked by the robot hand internal controller. 
Faster and more dynamic grasps require a proper motion control design, which we will pursue through adaptive feedback and tactile sensing.
In a similar fashion, object-template matching is based on simple segmentation and primitive fitting but could be significantly extended through learning mechanisms, allowing for a richer template library, including a  library whose templates could be learned through visuo-tactile sensing. 









\bibliographystyle{IEEEtran}
\bibliography{bibliography_v2_v3}

\end{document}